\documentclass[twoside]{article}
\RequirePackage[loading]{tracefnt}
\usepackage[T1]{fontenc}
\usepackage{lmodern}
\usepackage{xcolor}
\usepackage{amsmath}
\usepackage{amssymb}
\usepackage{mathtools}
\usepackage{amsthm}
\newcommand{\bm}[1]{\mathbf{#1}}
\usepackage{thm-restate}
\usepackage{algorithm}
\usepackage{algorithmic}
\usepackage{enumitem}
\usepackage{graphicx}
\usepackage{subcaption}
\usepackage{url} 

\usepackage[accepted]{aistats2025}
\usepackage[round]{natbib}

\theoremstyle{plain}
\newtheorem{theorem}{Theorem}[section]

\newtheorem{lemma}[theorem]{Lemma}

\newtheorem{corollary}[theorem]{Corollary}
\theoremstyle{definition}

\newtheorem{assumption}[theorem]{Assumption}
\theoremstyle{remark}

\begin{document}

%

%

\twocolumn[

\aistatstitle{Time-Varying Gaussian Process Bandits with Unknown Prior}

\aistatsauthor{ Juliusz Ziomek$^{1, *}$ \And Masaki Adachi$^{1,2}$ \And Michael A. Osborne$^1$}

\aistatsaddress{\\
$^1$Machine Learning Research Group, University of Oxford\\
$^2$Lattice Lab, Toyota Motor Corporation\\
$^*$ Corresponding Author, Correspondence to juliusz (dot) ziomek [at] univ (dot) ox.ac.uk
} ]

\begin{abstract}
  Bayesian optimisation requires fitting a Gaussian process model, which in turn requires specifying prior on the unknown black-box function---most of the theoretical literature assumes this prior is known. However, it is common to have more than one possible prior for a given black-box function, for example suggested by domain experts with differing opinions. In some cases, the type-II maximum likelihood estimator for selecting prior enjoys the consistency guarantee, but it does not universally apply to all types of priors. If the problem is stationary, one could rely on the Regret Balancing scheme to conduct the optimisation, but in the case of time-varying problems, such a scheme cannot be used.  To address this gap in existing research, we propose a novel algorithm, PE-GP-UCB, which is capable of solving time-varying Bayesian optimisation problems even without the exact knowledge of the function's prior.  The algorithm relies on the fact that either the observed function values are consistent with some of the priors, in which case it is easy to reject the wrong priors, or the observations are consistent with all candidate priors, in which case it does not matter which prior our model relies on. We provide a regret bound on the proposed algorithm. Finally, we empirically evaluate our algorithm on toy and real-world time-varying problems and show that it outperforms the maximum likelihood estimator, fully Bayesian treatment of unknown prior and Regret Balancing.
\end{abstract}

\section{INTRODUCTION}
\label{intro}
Bayesian Optimisation (BO; \citet{garnett2023bayesian}) based on a Gaussian Process (GP) emerged as a successful paradigm for solving black-box optimisation problems in a sample-efficient manner. In many practical application, such as controller design \citep{brunzema2022controller}, environment monitoring \citep{gao2022bayesian} or tuning hyperparameters of Reinforcement Learning algorithms \citep{parker2020provably}, the underlying black-box function will be changing over time. For such cases, a family of time-varying BO methods \citep{bogunovic2016time, nyikosa2018bayesian} have been developed. To conduct optimisation, BO algorithms first construct a surrogate model of the unknown black-box function based on the observed values for the points queried so far. GP is typically used as the model, due to its data efficiency, coming from assuming a prior on the unknown black-box function. 

This prior, however, needs to be specified by the user and its correct choice is paramount to the accurate predictions of the model. Such prior might be given to us by a domain expert, but as different experts might disagree, we might end up with a number of candidate priors. The seminal GP-UCB algorithm is proven \citep{srinivas2009gaussian} to enjoy a regret guarantee, but only if the function was sampled from a prior that we know. Using a different prior will produce a wrong posterior, which might cause the algorithm to select suboptimal points, resulting in high regret. 

Selecting prior by the maximising type-II likelihood \citep{good1983good, berger1985statistical} can be a valid strategy in certain scenarios and statistical literature provides some asymptotic guarantees \citep{kaufman2013role,li2022bayesian} as the number of observations increases, but they do not apply to all priors, particularly non-stationary ones. To universally guarantee that type-II MLE will recover the true prior (the prior black-box function was sampled from), one needs to cover the function domain with a dense, uniform grid \citep{bachoc2013cross}. However, this defeats the purpose of sample efficient optimisation, where instead of querying the function everywhere, we wish to use our limited query budget in the region close to the optimal solution. We would thus want to have an algorithm, which conducts optimisation while simultaneously trying out different priors and selecting the most promising one.

If the black-box function and search space do not change over time, one may employ the Regret Balancing \citep{pacchiano2020regret, ziomek2024bayesian} technique as a solution. Such a technique tries out multiple priors, analyses the function values obtained while using each of them and eliminates the underperforming ones. The regret of such a procedure is guaranteed to be `close' to the regret of an algorithm using the best prior among the candidates. However, if the problem exhibits any form of time-varying behaviour, the difference in function values observed while using different priors does not necessarily be due to the choice of prior, but rather to changes in the underlying problem. 

In this work, we propose the first GP-based BO algorithm which admits a provable regret bound under time-varying settings with an unknown prior. Our algorithm effectively eliminates the priors that are highly implausible given the data observed so far and when selecting the next point to query, remains optimistic with respect to the priors that have not yet been eliminated. Our proof technique is novel and relies on the idea that if all priors produce posteriors explaining the observations well enough then it does not matter which is used. On the other hand, if this is not the case then we can easily eliminate wrong models. We also conduct experiments and show PE-GP-UCB outperforms MLE, Fully Bayesian treatment of the unknown prior as well as the Regret balancing scheme.

\textbf{Related Work on time-varying GP bandits} The early work on GP bandits \citep{osborne2009gaussian,srinivas2009gaussian} considered stationary functions with search spaces constant throughout the optimisation process. The later work of \cite{bogunovic2016time} considered a setting where the function is evolving through time according to a probabilistic model. For this setting, they first derived posterior equations that work with time-varying data and then proposed a Time-Varying GP-UCB (TV-GP-UCB) algorithm, utilising those equations to conduct optimisation. However, their work assumes that the prior function and the probabilistic model of its evolution through time are known. Within this work, we lift this assumption and consider the case, where we have access to multiple candidate models, without the knowledge of which one is correct.

An orthogonal stream of research did not make any probabilistic assumptions on the black-box function and its evolution and instead assumed the function is a member of some Reproducing Kernel Hilbert Space \citep{williams2006gaussian}. The evolution of the black-box function was then constrained by a budget, which was assumed to be known. \citet{zhou2021no} proposed Resetting GP-UCB and Sliding Window GP-UCB algorithms, which respectively periodically reset the model or only use some number of most recent observations.  \citet{deng2022weighted} considered approaching this problem via a weighted GP model, where more weight is put on more recent observations.

\textbf{Related Work on BO with Misspecification}
The work of \cite{bogunovic2021misspecified} studied BO under model misspecification from the frequentist perspective, however, provided no method for finding the well-specified model. \cite{hvarfner2023self} proposed an algorithm that combines BO with active learning so that prior hyperparameters can be learnt simultanously while conducting optimisation. However, their method lacks theoretical guarantees and introduces substantial computational overhead. \cite{rodemann2024imprecise} proposed an algorithm to conduct BO when instead of one prior mean function, we are provided a set of possible mean functions, however, provided no regret guarantees for this algorithm. \cite{xu2024principled} proposed an algorithm for conducting BO with preferential feedback, where the black-box function value itself cannot be directly evaluated. In a way, one can think about this problem settings as a misspecification of the observed function value. To address this authors propose to use likelihood based confidence sets and remain optimistic with respect to the best feasible function in the confidence set. However, this work assumes frequentist setting and that kernel hyperparameters are known.

\textbf{Related Work on Master Algorithms}  
Regret Balancing scheme \citep{abbasi2020regret, pacchiano2020regret} provides a way to create a master algorithm to coordinate a number of base algorithms for solving multi-armed bandit-style problems. Simiarly as done by \cite{ziomek2024bayesian}, one could technically introduce one GP-UCB algorithm for each prior as the set of base learners and then use Regret Balancing to aggregate them into one master algorithm, but as we explained before, its theoretical guarantees on optimality do not hold when the function or search space is time-varying. \cite{hoffman2011portfolio} proposed an algorithm GP-Hedge which uses an adversarial bandit \citep{lattimore2020bandit} strategy for selecting one of many acquisition functions. However, the regret bound they derived is incomplete, as they were not able to bound the sum of predictive variances at the points queried by the algorithm.

\section{PROBLEM STATEMENT}

We consider a problem setting where we wish to maximise an unknown, black-box and spatio-temporal function $f: \mathcal{X} \times [T]  \to \mathbb{R}$ over some compact set $\mathcal{X} \subset \mathbb{R}^d$.  We assume we are allowed to query a single point $\bm{x}_t$ at each timestep $t \in [T]$ and obtain its corrupted function value $y_t = f(\bm{x}_t, t) + \epsilon_t$ by some i.i.d. Gaussian noise  $\epsilon_t \sim \mathcal{N}(0, R^2)$ for all $t  \in [T]$. The function is allowed to change with time step $t$, hence the explicit dependence on $t$ in its signature. We also assume the available domain at each time step $\mathcal{X}_t \subseteq \mathcal{X}$ can change in an arbitrary manner, possibly chosen by an adversary. As such, formally, at each time step, the optimal point we can select is $\bm{x}^*_t = \arg \max_{\bm{x} \in \mathcal{X}_t} f(\bm{x}, t)$.
As we want to maximise the function, the instantaneous regret is defined as $r_t =  f(\bm{x}^*_t, t) - f(\bm{x}_t, t)$. Similarly, we define cumulative regret as $R_T = \sum_{t=1}^T r_t$. We assume that the black-box function and its evolution through time are governed by some prior, i.e., $f \sim p^\star(f)$ and we refer to $p^\star(\cdot)$ as the true prior. We also assume we are given access to a set of priors $\mathcal{U}$ provided to us (e.g. by domain experts) that contains the true prior $p^*(f)$. However, we also assume that we do not know, which prior in $\mathcal{U}$ is the true one.

Each GP prior $p(f)$ is uniquely defined by its spatio-temporal mean function $\mu^p(\cdot): \mathcal{X} \times [T] \to \mathbb{R}$ and kernel function $k^p(\cdot, \cdot): (\mathcal{X} \times [T]) \times (\mathcal{X} \times [T]) \to \mathbb{R} $. We require $\forall_{t \in [T]} \sup_{\bm{x}  \in \mathcal{X}} |\mu^p(\bm{x}, t)| < \infty$ for each $p \in 
\mathcal{U}$.  Without the loss of generality, we also assume that $k^p(\cdot, \cdot) \le 1$ for each $p \in \mathcal{U}$. An example of such spatio-temporal kernel could be the time-varying RBF kernel:
\begin{equation*}
    k((\bm{x},t), (\bm{x}^\prime,t^\prime)) = \exp\left(-\frac{1}{2 l}\lVert \bm{x} - \bm{x}^\prime  \rVert_2^2\right) (1-\varepsilon)^{|t - t^\prime| / 2},
\end{equation*}
where $l$ and $\varepsilon$ are hyperparameters. Note that one could think about spatio-temporal kernels as standard stationary kernels with the input space extended to include one additional variable $t$ \citep{van2012kernel, van2012estimation}. The only difference is that when conducting optimisation, we have no control over the choice of the $t$ variable, as opposed to the $\bm{x}$ variable.

As the difficulty of the problem depends heavily on the  kernel $k^{p^*}(\cdot,\cdot)$ of the prior $p^*(\cdot)$ the function was sampled from, typically in preceding literature, the regret bound is expressed in terms of maximum information gain (MIG, \cite{srinivas2009gaussian}) of kernel $k^{p}$, defined as $
\gamma_T^{p} := \max_{x_1, \dots, x_T} \frac{1}{2} \log \lvert I + R^{-2} \bm{K}^{p}_T \rvert$, where $\mathbf{K}^p_{T} \in \mathbb{R}^{T \times T}$ with entries $(\mathbf{K}^p_T)_{i,j} = k^p((\bm{x}_i, i), (\bm{x}_j, j))$. \cite{srinivas2009gaussian} derived bounds on MIG for standard kernels and then \cite{bogunovic2016time} extended these bounds to spatio-temporal kernels.

\textbf{Gaussian Process Model} In GP-based BO, at each timestep $t$, we contruct a surrogate model of the black-box function, based on the queries so far. Let $\mathcal{D}_{t-1} = \{(\bm{x}_i, y_i)\}_{i=1}^{t-1}$
be the collection of all queried points and their corresponding noisy values up to timestep $t-1$. If we use a GP model with a prior $p(\cdot)$ then given  the data so far $\mathcal{D}_{t-1} $, the model returns us the following mean $\mu^p_{t-1}(\bm{x}) $ and variance $(\sigma^p_{t-1})^2(\bm{x})$ functions:
\begin{align*}
    \mu^p_{t-1}(\bm{x}, t^\prime) =& (\bm{k}^p_{t-1}(\bm{x}, t^\prime)^T(\mathbf{K}^p_{t-1}+ R^2\mathbf{I})^{-1} \\
     &(\bm{y}_{t-1} - \mu^p(\bm{x}, t^\prime)) ) + \mu^p(\bm{x}, t^\prime)
\end{align*}
\begin{align*}
    (\sigma^p_{t-1})^2(\bm{x}, t^\prime) =& k^p((\bm{x}, t^\prime), (\bm{x}, t^\prime)) -\\
    & \bm{k}^p_{t-1}(\bm{x}, t^\prime)^T(\mathbf{K}^p_{t-1} + R^2\mathbf{I})^{-1}\bm{k}^p_{t-1}(\bm{x}, t^\prime)
\end{align*}
where $\bm{y}_{t-1} \in \mathbb{R}^{t-1} $ with elements $(\bm{y}_{t-1})_i = y_i$, $\bm{k}^p_{t-1}(\bm{x}, t^\prime) \in \mathbb{R}^{t-1}$ with elements $\bm{k}^p_{t-1}(\bm{x}, t^\prime)_i = k^p((\bm{x}, t^\prime), (\bm{x}_i, i))$. Note the distinction between $t$ and $t^\prime$. The former refers to the time step in optimisation, whereas the latter to the point in time for which we wish to make a prediction. While in the algorithm we will always use $t = t^\prime$, as we wish to make a prediction about current timestep, this distinction will be useful for proving one of suplementary Lemmas. While using mean and variance functions provided by the GP, we get the guarantee on predictive performance, as stated in the next Section.

\section{PRELIMINARIES}
We now recall well-known results from the literature. Those results would be later needed to prove the regret bound of our proposed algorithm. This results come from the work of \cite{bogunovic2016time}, which extend the stationary setting result from \cite{srinivas2009gaussian}.
\begin{theorem}[Appendix C1 of \cite{bogunovic2016time}] \label{theorem:ucb}
Let $f \sim p(f)$, and set $\beta^{p}_t = \sqrt{2\log\left( \frac{|\mathcal{X}| \pi^2 t^2}{2\delta_A} \right)}$. Then, with probability at least $1 - \delta_A$, for all $\bm{x} \in \mathcal{X}$ and all $t \in [T]$:
    \begin{equation*}
    \left|f(\bm{x}, t) -  \mu_{t-1}^{p}(\bm{x}, t)\right| \le \beta^{p}_{t} \sigma_{t-1}^{p} (\bm{x}, t) .
\end{equation*}
\end{theorem}

 Note that the confidence parameter $\beta_t^p$ in Theorem \ref{theorem:ucb}, depends on $|\mathcal{X}|$ in the Bayesian setting and only makes sense if $|\mathcal{X}| < \infty$. If this is not the case, we need to adopt a standard assumption \citep{srinivas2009gaussian, bogunovic2016time}.
  \begin{assumption} \label{as:lkernel}
    Let $X \subset [0, r]^d$ be a compact and convex set, where $r > 0$. For each prior $p \in \mathcal{U}$, assume that the kernel $k^{p}$ satisfies the following condition on the derivatives of a sample path $f \sim p(f)$. There exist constants $a^p, b^p > 0$, such that:
\[
P \left(
\sup_{x \in \mathcal{X}} \left\lvert \frac{\partial f}{\partial x_j} \right\rvert  > L
\right)
\leq a^p \exp \left( - \frac{L^2}{b^p} \right),
\]
for each $j \in [d]$.
\end{assumption}
 Under that assumption, we can derive high-probability confidence intervals even if $|\mathcal{X}| = \infty$.  Following \citep{srinivas2009gaussian, bogunovic2016time}, purely for the sake of analysis, let us define a discretization $D_t$ of size $(\tau_t)^d$ so that for all $x \in D_t$, $\lVert x - [x]_t \rVert_1 \leq \frac{r}{d\tau_t}$, where $[x]_t$ denotes the closest point in $D_t$ to $x$. We now get the following Theorem.

\begin{theorem}[Appendix C1 of \cite{bogunovic2016time}] \label{theorem:ucb_bayesian_infinite}
Let Assumption \ref{as:lkernel} hold, set
$\beta_t^{p} = \sqrt{2 \log\left(\frac{\pi^2 t^2}{2\delta_A}\right) + 2d \log\left( dt^2rb^p\sqrt{\log\left(\frac{da^p}{2\delta_A}\right)}\right)}
$ and set the size of discretisation to $\tau_t = dt^2 rb^p\sqrt{\log\left(\frac{2da^p}{\delta_A}\right)}$. Then  with probability at least $1 - \delta_A$ we have $\forall \bm{x} \in \mathcal{X}$ and $\forall t \in [T]$: 
\begin{equation*}
   |f(\bm{x}, t) - \mu_{t-1}^{p}( [\bm{x}]_t, t)| \le \beta_{t}^{p} \sigma^{p}_{t-1}([\bm{x}]_t, t) + \frac{1}{t^2} .
\end{equation*}
And for any $\bm{x}_t$ deterministic conditioned on $\mathcal{D}_{t-1}$ :
\begin{equation*}
    |f(\bm{x}_t, t) - \mu_{t-1}^{p}(\bm{x}_t, t)| \le \beta_{t}^{p} \sigma^{p}_{t-1}(\bm{x}_t, t).
\end{equation*}

\end{theorem}

\section{GP-UCB WITH PRIOR ELIMINATION}

\begin{figure}
    \centering
    \includegraphics[width=\hsize]{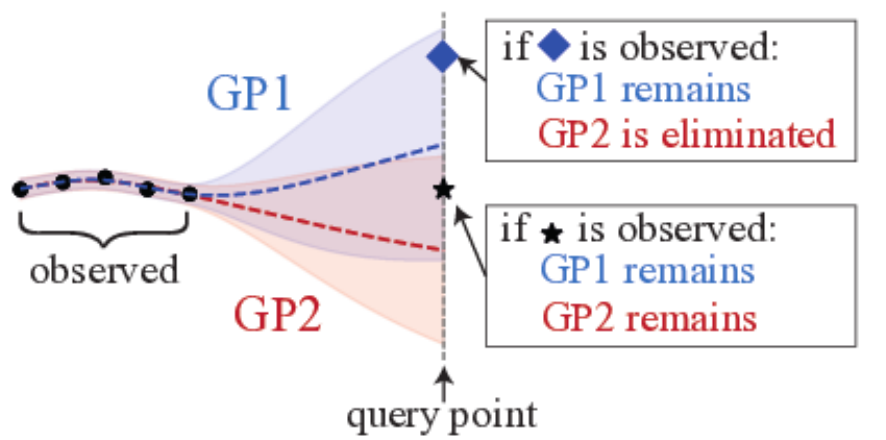}
    \caption{Intuition behind the prior elimination. Solid lines represent the mean functions of two models - GP1 and GP2, produced by fitting the observed points while assuming different prior, and shaded regions are confidence intervals. The point marked by the star lies in the confidence intervals of both models, whereas the point marked by the diamond lies in the one of GP1.}
    \label{fig:different_ucb}
\end{figure}

We now state our proposed algorithm in Algorithm \ref{alg:unknown_prior_bo_noisy}. When selecting the next point to query (line 7), our algorithm is doubly optimistic, with respect to both the function value and the model. In other words, for each point, we use the largest possible upper confidence bound among the ones suggested by models defined by possible priors in $
\mathcal{U}_{t-1}$. We denote the selected point by $\bm{x}_t$ and the prior under which it achieved the highest UCB value is denoted by $p_t(\cdot)$. The set $S^p_t$ stores all iterations $i$ up to the current one $t$, during which $p_t(\cdot) = p(\cdot)$, for each $p \in \mathcal{U}$ (line 8). After observing the function value, we calculate the error $\eta_t$ between the observed value and the mean at point $\bm{x}_t$, according to the model defined by $p_t(\cdot)$ (line 11). Then, each model for which the sum of errors is too big is removed from the set of candidates (lines 12-14) and we continue with the loop. As we will show later, the true prior $p^\star$ will never be rejected with high probability. As such, if all priors have been rejected, this likely indicates that none of the supplied priors was the true prior. In such case, the algorithm could terminate and flag this issue to the user.

The intuition behind the rejection mechanism is depicted in Figure \ref{fig:different_ucb}.  Let us assume we have two models GP1 and GP2, represented by red and blue solid lines, together with their corresponding confidence intervals in dashed lines. After we query a new point, it can either lie within the confidence intervals of both models (like the star) or only one model (like the diamond). If we were to obtain a point like the diamond, we clearly see that the red model's prediction was very inaccurate and we lean towards rejecting the prior that produced it. On the other hand, if we observe a point like the star when querying the function then it does not matter, which model is the "true" one, as such an observation is probable under both models and the errors for either of the models would be small. Thanks to this mechanism, we can conduct BO, whilst simultaneously eliminating the (most-likely) wrong priors. For such a strategy, we can derive a regret bound, as stated by the following Theorem.

\begin{algorithm}
\caption{Prior Elimination GP-UCB \\(PE-GP-UCB) }\label{alg:unknown_prior_bo_noisy}
\begin{algorithmic}[1]
\REQUIRE a set of possible priors  $\mathcal{U}$; confidence parameters $\{\xi_t\}_{t=1}^T$ and
$\{\{\beta^p_t\}_{t=1}^T\}_{p \in \mathcal{U}}$ 

\STATE Set $\mathcal{D}_{0} = \emptyset$, $\mathcal{U}_0 = \mathcal{U}$, $S^{p} = \emptyset$ for each $p \in \mathcal{U}$
\FOR{$t = 1, \dots, T$}
\FOR{$p \in \mathcal{U}_{t-1}$}
\STATE Fit a GP $( \mu^p_{t-1}(\bm{x}, t), \sigma^p_{t-1}(\bm{x}, t))$ to $\mathcal{D}_{t-1}$ 
\STATE Set $\textrm{UCB}_{t-1}^p(\bm{x}) = \mu^p_{t-1}(\bm{x}, t) + \beta_t^p \sigma^p_{t-1}(\bm{x}, t)$
\ENDFOR
\STATE Find $\bm{x}_t, p_t =  \underset{\bm{x},  p  \in \mathcal{X} \times \mathcal{U}_{t-1}}{\arg\max} \textrm{UCB}_{t-1}^p(\bm{x})$
\STATE $S^{p_t}_t = S^{p_t}_{t-1} \cup \{t\}$ and $S^{p}_t = S^{p}_{t-1} $ for $p \in \mathcal{U}; p\neq p_t$
\STATE Query the black-box $y_t = f(x_t)$ 
\STATE Set $\mathcal{D}_t = \mathcal{D}_{t-1} \cup (x_t, y_t)$
\STATE Set $\eta_t =  y_t - \mu^{p_t}_{t-1}(\bm{x}_t)$ 
\IF{$\left | \sum_{i \in S^{p_t}_t} \eta_i \right| > \sqrt{\xi_t |S^{p_t}_t|} + \sum_{i \in S^{p_t}_t} \beta^{p_t}_i \sigma_i^{p_t}(\bm{x}_i)$} 
\STATE $\mathcal{U}_t = \mathcal{U}_{t-1} \setminus \{p_t\}$ 
\ENDIF
\ENDFOR

\end{algorithmic}
\end{algorithm}

\begin{restatable}[]{theorem}{noregretbayesian}
\label{theorem:noregret} Let $f \sim p^*(f)$, $\epsilon_t \sim \mathcal{N}(0, R^2)$ for each $t$ and  $p^* \in \mathcal{U}$ and denote $\gamma_T = \max_{p \in \mathcal{U}} \gamma_T^p$. Set $\xi_t = 2 R^2 \log \frac{|\mathcal{U}| \pi^2 t^2}{\delta}$ and :
\begin{enumerate}[label=(\Alph*)]
    \item if $|\mathcal{X}| < \infty$, set $\beta_t^p = \sqrt{2\log \frac{2 |\mathcal{X}| \pi^2 t^2}{\delta}}$
    \item if  $|\mathcal{X}| = \infty$, let Assumption \ref{as:lkernel} hold and  set
    
\end{enumerate}
{\small $$\beta_t^{p} =  \sqrt{2 \log\left(\frac{3\pi^2 t^2}{2\delta}\right) + 2d \log\left( dt^2rb^p\sqrt{\log\left(\frac{3da^p}{2\delta}\right)}\right)}
$$}
 Then, Algorithm \ref{alg:unknown_prior_bo_noisy} with probability at least $1 - \delta$ achieves the cumulative regret of at most:
    \begin{equation*}
        R_T \le  \mathcal{O} \left( |\mathcal{U}|B_{t^\star} + \sqrt{|\mathcal{U}|\xi_T T} +  \beta_T \sqrt{|\mathcal{U}|T \gamma}_{T} \right)
    \end{equation*}
     where $\beta_T = \max_{p \in \mathcal{U}} \beta_T^p$, $\gamma_T = \max_{p \in \mathcal{U}} \gamma_T^p$ and $t^\star$ is the last iteration where any prior was rejected and $B_{t^\star} = \beta_t^{p^\star} +  \max_{t^\prime \in [t^\star]} \sup_{\bm{x} \in \mathcal{X}}|\mu^{p^\star}(\bm{x}, t^\prime)| + 1$.
\end{restatable}

\section{PROOF OF THE REGRET BOUND}
    We now provide proof of the regret bound. We  focus on the case of $|\mathcal{X}| < \infty$ and then comment on differences when $|\mathcal{X}| = \infty$.
    We first introduce the following Lemma for the concentration of noise, proved in Supplementary Material Section \ref{ap:noisebound}.
    
    \begin{restatable}[]{lemma}{noisebound}
\label{lemma:noisebound}
For each $p \in \mathcal{U}$ and $t\in [T]$ we have:
\begin{equation*}
    P \left( \underset{t= 1,\dots, T}{\forall} \, \underset{p \in \mathcal{U}}{\forall} \left | \sum_{i \in S_t^p} \epsilon_i \right| \le \sqrt{\xi_t |S_t^p|} \right) \ge 1 - \delta_B ,
\end{equation*}
 where $\xi_t = 2 R^2 \log \frac{|\mathcal{U}|\pi^2 t^2}{3\delta_B}$.
\end{restatable}

We also introduce the following result, which allows us to bound the maximum absolute value 
of the function that could be sampled from the prior.
\begin{restatable}[]{lemma}{bayesianfunctionbound}
\label{lemma:bayesianfunctionbound}
For $f \sim p(f)$ with some mean and kernel functions $\mu^p(\cdot)$ and $k^p(\cdot, \cdot)$ with properties $|\mu^p(\cdot)| < \infty $ and $k^p(\cdot, \cdot) \le 1$, for any $t \le T $ with probability at least $1 - \delta_C$ we have:
\begin{equation*}
      \max_{t^\prime \in [t]}\sup_{\bm{x} \in \mathcal{X}} \left| f(\bm{x}, t^\prime) \right|\le  B_t,
\end{equation*}
where $B_t =  \beta_t^{p^\star} + \max_{t^\prime \in [t]} \sup_{\bm{x} \in \mathcal{X}}|\mu^{p^\star}(\bm{x}, t^\prime)| + 1 $ and $\beta_T^{p^\star}$ is to be set as in Theorem \ref{theorem:ucb} for discrete case and as in Theorem \ref{theorem:ucb_bayesian_infinite} for continuous case with $\delta_A$ replaced with $\delta_C$.

\end{restatable}
Observe that by union bound Theorem \ref{theorem:ucb} and Lemma \ref{lemma:noisebound} and Lemma \ref{lemma:bayesianfunctionbound}) hold together with probability at least $1 - \delta_A - \delta_B - \delta_C$. We choose $\delta_A = \delta_B =\delta_C = \frac{\delta}{3}$. We prove the regret bound, assuming the probabilistic statements in Theorem \ref{theorem:ucb} and Lemma \ref{lemma:noisebound} hold, and as such the resulting bound holds with probability at least $1-\delta$.

\textbf{Preservation of $p^*(\cdot)$} We first show that if the statements from  Theorem \ref{theorem:ucb} and Lemma \ref{lemma:noisebound} hold, then the true prior  $p^*(\cdot)$ is never rejected. Indeed, when line 12 is executed at iteration $t$: 
\begin{equation*}
       \left| \sum_{i \in S^{p^*}_t} \eta_i \right|  
    \end{equation*}
    \begin{equation*}
        =   \left| \sum_{i \in S^{p^*}_t} \left( y_i - f(\bm{x}_i, i) + f(\bm{x}_i, i) - \mu^{p^*}_{i-1}(\bm{x}_i, i) \right) \right|
    \end{equation*}
    \begin{equation*}
        \le \left | \sum_{i \in S^{p^*}_t} \epsilon_i \right| +  \sum_{i \in S^{p^*}_t}  \left |  f(\bm{x}_i, i) - \mu^{p^*}_{i-1}(\bm{x}_i, i)\right|  
    \end{equation*}
    
    \begin{equation*}
        \le \sqrt{\xi_t |S^{p^*}_t|} + \sum_{i \in S^{p^*}_t} \beta_i^{p^*}\sigma_i^{p^*}(\bm{x}_i, i)  ,
    \end{equation*}
where the last inequality is due to Theorem \ref{theorem:ucb} and Lemma \ref{lemma:noisebound}. As such, the condition of the \texttt{if} statement always evaluates to \texttt{false} and  $p^* \in \mathcal{U}_{t}$ for any $t \ge 0$. Next, we show how this fact allows us to bound instantaneous regret.

\textbf{Bound on simple regret} If $p^* \in \mathcal{U}_{t-1}$  we trivially have:
    \begin{align} 
        f(\bm{x}^*_t,t) & \le \mu_{t-1}^{p^*}(\bm{x}^*_t,t) + \beta_t^{p^*} \sigma^{p^*}_{t-1}(\bm{x}^*_t,t) \\
         & \le \max_{\bm{x}_t \in \mathcal{X}_t} \max_{p \in \mathcal{U}_t} \mu_{t-1}^p(\bm{x}_t, t) + \beta_t^{p} \sigma_{t-1}^p(\bm{x}_t, t), \label{eq:uucb_maintext}
    \end{align}
    where first inequality is due to Theorem \ref{theorem:ucb}.
     For simplicity we will write $\mu_{t-1}^+ = \mu_{t-1}^{p_{t}}$, $\sigma_{t-1}^+ = \sigma_{t-1}^{p_{t}}$ and $\beta^+_t = \beta_t^{p_t}$. 
     Having established an upper bound on $f(\bm{x}^*_t, t)$, it remains to find a lower bound on $f(\bm{x}_t, t)$. We have also that:
     \begin{equation} \label{eq:llcb_maintext}
         f(\bm{x}_t, t) = y_t - \epsilon_t =  \mu^+_{t-1}(\bm{x}_t, t)  + \eta_t - \epsilon_t .
     \end{equation}
     The observation above is a key component of our proof, as it relates the function value to the prediction error of $\eta_t$ of the model defined by prior $p_t$. Combining (\ref{eq:uucb_maintext})-(\ref{eq:llcb_maintext}) yields the following bound on instantaneous regret:
    \begin{equation} \label{eq:instant_regret_bound}
        r_t = f(\bm{x}^*_t, t) - f(\bm{x}_t, t)  \le  \beta_t^+ \sigma^+_{t-1}(\bm{x}_t, t)  - \eta_t + \epsilon_t .
    \end{equation}

 \textbf{Bound on cumulative regret} 
    With our bound on instantaneous regret, we now proceed to bound the cumulative regret. Let us define $\mathcal{C}$ to be the set of `critical' iterations  as below:
{\small \begin{equation*} 
        \mathcal{C} = \left \{t \in [T]: \left | \sum_{i \in S^{p_t}} \eta_i \right| > \sqrt{\xi_t |S^{p_t}|} + \sum_{i \in S^{p_i}} \beta^{p_i}_i \sigma_i^{p_t}(\bm{x}_i, i) \right\} 
\end{equation*}}
Notice that for each $t \in \mathcal{C}$, we discard one possible prior and as such $|\mathcal{C}| \le |\mathcal{U}|$. On those iterations, we may suffer an arbitrarily high regret. Due to Lemma \ref{lemma:bayesianfunctionbound} we have $r_t \le 2B_t$ for all $t \in \mathcal{C}$.  At each timestep, we can reject a prior and expand $\mathcal{C}$ or not reject any of the priors and then $\mathcal{C}$ does not grow. As there is finite number of priors we must have that $|\mathcal{C}| < |\mathcal{U}|$ and there must be some final iteration $t^\star$ after which $\mathcal{C}$ was never expanded anymore. As $B_t$ can only grow with $t$, we have that $\forall_{t \in \mathcal{C}} B_t \le B_{t^\star}$ and thus $\forall_{t \in \mathcal{C}} r_t \le 2B_{t^\star}$. Using Equation \ref{eq:instant_regret_bound}, we thus see that the cumulative regret is bounded as:
\begin{equation*}
    R_T \le \sum_{t \in \mathcal{C}} r_t + \sum_{t \notin \mathcal{C}} r_t 
\end{equation*}
\begin{equation*}
    \le 2|\mathcal{U}|B_{t^\star} +  \sum_{t \notin \mathcal{C}}\beta^+_t \sigma^{+}_{t-1}(\bm{x}_t, t) + \sum_{p \in \mathcal{U}} \sum_{t \in S^u_T \setminus \mathcal{C}}  \left(  \epsilon_t - \eta_t  \right)  .
\end{equation*}

 When it comes to the terms $\eta_t$, observe that if $t \notin \mathcal{C}$ the \texttt{if} statement in line 12 evaluates to \texttt{false}. Thus:
{ \small
 \begin{equation*}
     \sum_{p \in \mathcal{U}}\sum_{t \in S^u_T \setminus \mathcal{C}} - \eta_t \le \sum_{p \in \mathcal{U}}\sqrt{\xi_T |S^{p}_T|} + \sum_{p \in \mathcal{U}}\sum_{t \in S^u_T \setminus \mathcal{C}}  \beta^{p}_t \sigma_{t-1}^{p}(\bm{x}_t, t).
 \end{equation*}}
 We combine this fact together with Lemma \ref{lemma:noisebound} to arrive at the following:
\begin{align*}
        R_T &\le 2|\mathcal{U}|B_{t^\star} + 2 \sum_{p \in \mathcal{U}}\sqrt{\xi_T |S^p_T|}  + 2 \sum_{t\notin \mathcal{C}} \beta_t^+ \sigma^+_{t-1}(\bm{x}_t, t) \\
        & \le 2|\mathcal{U}|B_{t^\star} + 2 \sqrt{\xi_T|\mathcal{U}|T} + 2 \sum_{t\notin \mathcal{C}} \beta_t^+ \sigma^+_{t-1}(\bm{x}_t, t),
    \end{align*}
where the last inequality is due to Cauchy-Schwarz.

\textbf{Expressing bound in terms of MIG} The final step of the proof is to express the third term of the cumulative regret bound in terms of maximum information gain. We do this in the following Lemma, proved in the Supplementary Material.

 \begin{restatable}[]{lemma}{maxinfogain}
\label{lemma:maxinfogain}
There exists a constant $C > 0$, such that:
\begin{equation*}
      \sum_{t\notin \mathcal{C}} \beta^+_t \sigma^+_{t-1}(\bm{x}_t, t) \le \beta_T \sqrt{CT |\mathcal{U}|  \gamma_{T}} ,
\end{equation*}
where $\beta_T = \max_{p \in \mathcal{U}} \beta_T^p$ and $\gamma_T = \max_{p \in \mathcal{U}} \gamma_T^p$.
\end{restatable}

    Applying Lemma \ref{lemma:maxinfogain} to previously developed cumulative regret bound, we get:
    \begin{equation*}
         R_T \le 2|\mathcal{U}| B_{t^\star}+ 2 \sqrt{\xi_T|\mathcal{U}|T} + 2\beta_T \sqrt{CT |\mathcal{U}|  \gamma_{T}},
    \end{equation*}
which finishes the finite $\mathcal{X}$ case bound.

\textbf{Differences in Continous Case}
If $|\mathcal{X}| = \infty$, then instead of Theorem \ref{theorem:ucb}, we rely on Theorem \ref{theorem:ucb_bayesian_infinite} and the rest of the proof works in the same way, except for the fact that the instantaneous regret has an additional $\frac{1}{t^2}$ term. However, as $\sum_{t=1}^T \frac{1}{t^2} \le \frac{\pi}{6}$, it just adds a constant that does not affect the scaling of the bound.

\section{DISCUSSION}
The first term in our bound $|\mathcal{U}|B_{t^\star}$ depends on $B_{t^\star} = \beta_1 +  \max_{t^\prime \in [t^\star]} \sup_{\bm{x} \in \mathcal{X}}|\mu^{p^\star}(\bm{x}, t^\prime)| + 1$. Note that as there is a finite number of priors, we cannot reject priors infinitely, as stated by next Lemma.
\begin{lemma} \label{lemma:tstar}
    We have that $P(\lim_{T \to \infty} \frac{t^\star}{T} \to 0) = 1$ and $P(\lim_{T \to \infty} \frac{B_{t^\star}}{T} \to 0) = 1$.
\end{lemma}

As such for sufficiently large $T$, $t^\star$ will not depend on $T$ and as such $B_{t^\star}$ will also not depend on $T$ and the first term will get dominated by the remaining terms.   As for most commonly used kernels, the MIG $\gamma_T$ is at least logarithmic in $T$  \citep{srinivas2009gaussian}, the third term dominates the second term. In case where prior is known, GP-UCB (and its time-varying variant) admits a regret bound of $R_T \le \mathcal{O}(\beta_T \sqrt{T \gamma_T^{p^\star}})$ \citep{srinivas2009gaussian, bogunovic2016time}. As such, our bound is $\sqrt{|U|}$ factor away and scales with $\gamma_T = \max_{p \in \mathcal{U}} \gamma_T^p$ instead of $\gamma^{p^\star}_T$. This is an artifact of the fact that PE-GP-UCB is doubly-optimistic and if one prior produces upper confidence bound that is always the highest among other priors, it will always be selected as $p_t$. As such, PE-GP-UCB cannot be used when priors differ only with respect to outputscale.  However, in many applications, $\gamma^p_T$ will be the same for all priors $p \in \mathcal{U}$, for example, in the case where priors differ with respect to mean only, and then $\gamma_T = \gamma_T^{p^\star}$. Even if this is not the case, suffering regret scaling with  $\gamma_T $ is still much better than the arbitrarily high regret we may suffer if we rely on the wrong prior, which might happen if the prior is chosen by an alternative method without a correctness guarantee. 

The scaling with the square root of the number of possible priors $\sqrt{|U|}$ can be problematic when $|U|$ is large, but this rate is actually better than for Regret Balancing \citep{pacchiano2020regret, abbasi2020regret}, where the time-dependent part of bound grows linearly with number of possible models. In general, linear dependence on the number of base algorithms is common among Master Algorithms literature, where, for example, the celebrated CORRAL algorithm \citep{agarwal2017corralling} also achieves at least linear scaling. As such, we believe the square root scaling $\sqrt{|U|}$ is relatively favourable, when compared to algorithms solving similar problems.

Our bound can be applied to different types of kernels, given that their MIG is known. For concreteness, we now plug the expression from \citep{bogunovic2016time} for MIG of two commonly used time-varying kernels.

\begin{corollary} \label{corollary:win}
Under the same conditions as Theorem \ref{theorem:noregret}, the following statements hold:
    \begin{itemize}
        \item Let all candidate priors utilise the squared exponential kernel, we then have that  regret of PE-GP-UCB: 
        $R_T = \Tilde{\mathcal{O}}( |\mathcal{U}| B_{t^\star} + \sqrt{|\mathcal{U}|}\max\{T\varepsilon^{1/6}, \sqrt{T}\})$
        \item Let all candidate priors utilise the $\nu$-Matern kernel, we then have that regret of PE-GP-UCB:
        $R_T = \Tilde{\mathcal{O}}(|\mathcal{U}| B_{t^\star} + \sqrt{|\mathcal{U}|}\max\{T\varepsilon^{\frac{1}{2} \frac{1 - c}{3 - c}}, \sqrt{T^{1+c}}\})$
    \end{itemize}
\end{corollary}

\begin{figure*}[ht]
    \centering
    \begin{subfigure}[b]{0.49\textwidth}
        \includegraphics[width=\textwidth]{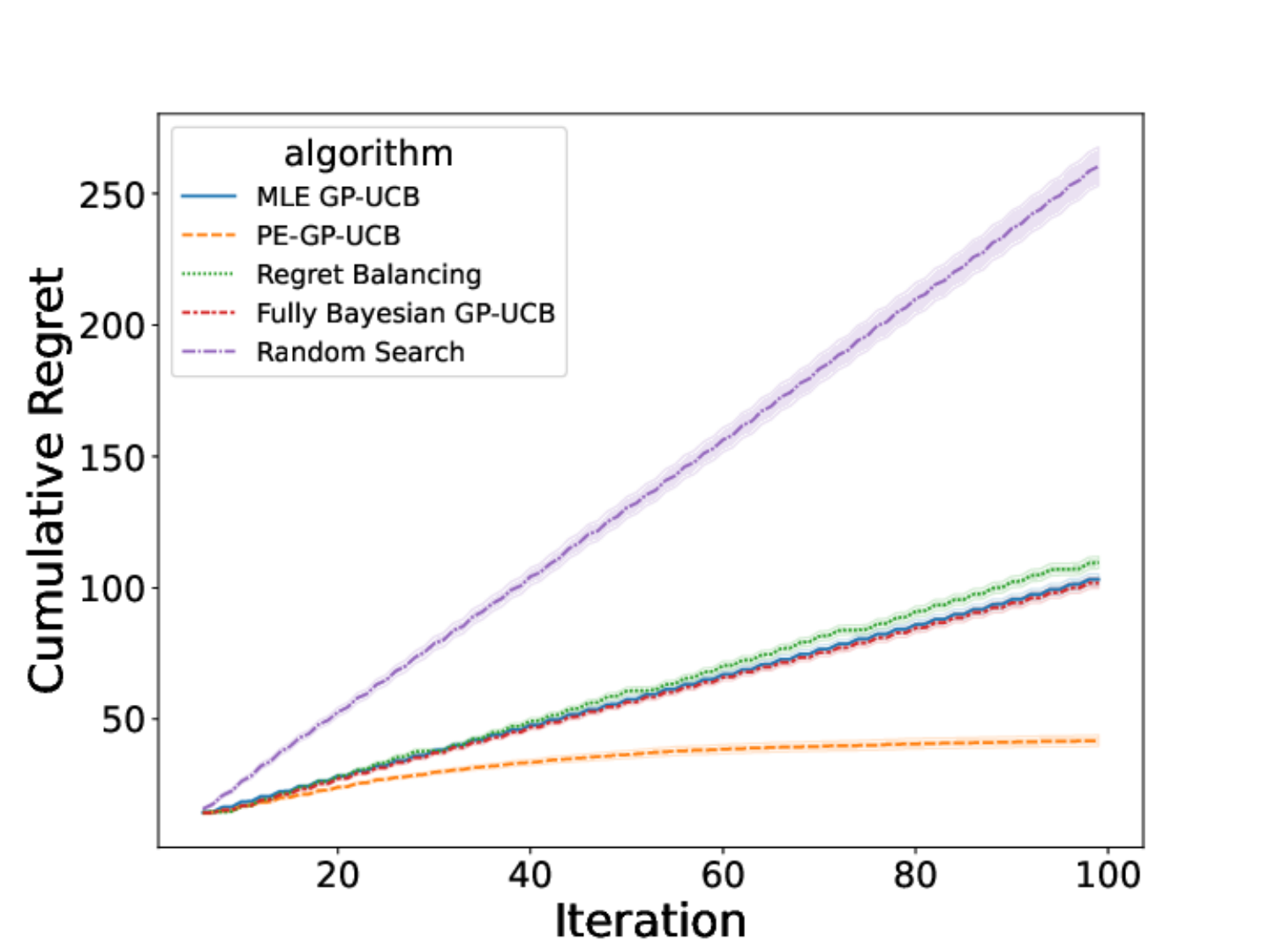}
        \caption{Regret results.}
        \label{fig:advregret}
    \end{subfigure}
    \hfill
    \begin{subfigure}[b]{0.49\textwidth}
        \includegraphics[width=\textwidth]{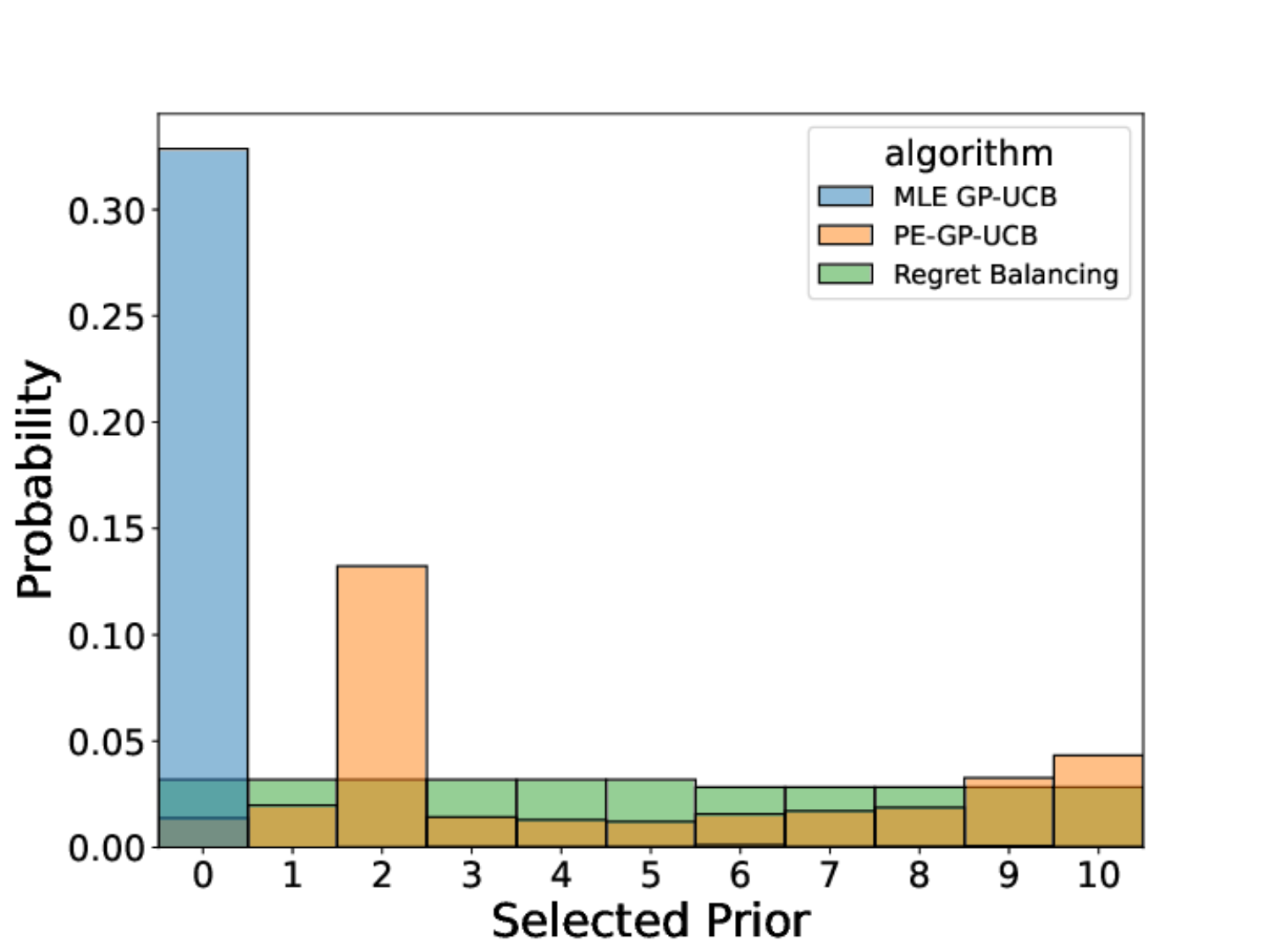}
        \caption{Histogram of prior selection frequency.}
        \label{fig:advhist}
    \end{subfigure}
    \caption{Results on the toy problem. We ran 30 seeds and in subfigure a) we show the mean values in solid lines and standard errors as shaded regions. In subfigure b) we show the proportion of times that the methods chose each prior accross all timesteps and all seeds. The $n$th prior has the $n$th hill tallest and the prior number $0$ has all hills of equal height.}
    \label{fig:toy_problem}
\end{figure*}

\begin{figure}[h]
    \centering
    \includegraphics[width=\linewidth]{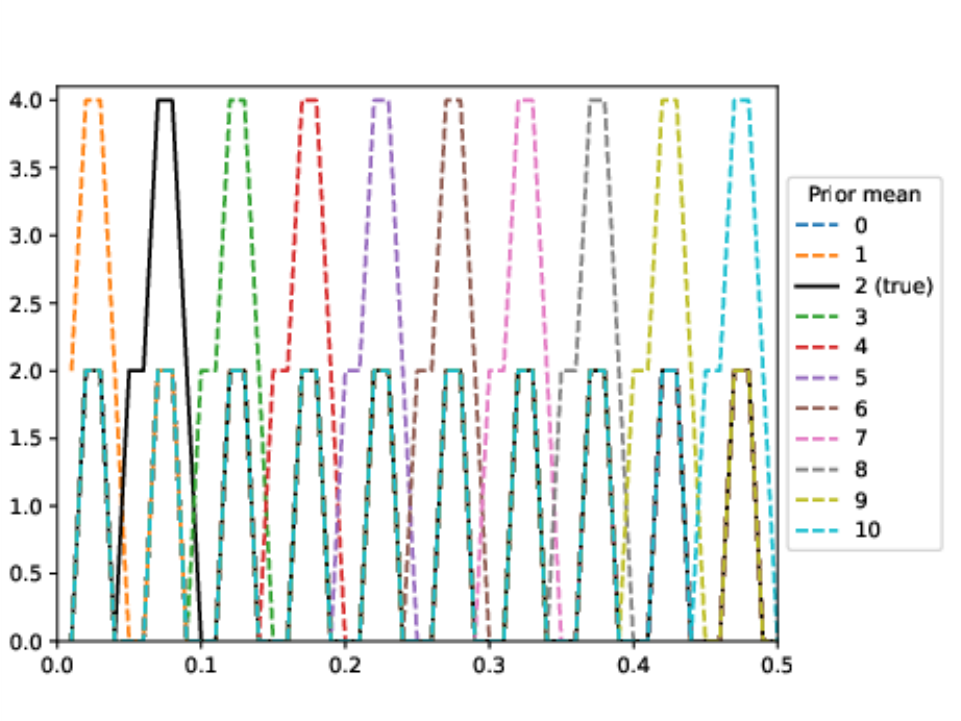}
    \caption{Plot of mean functions of different priors used for the toy problem. All priors used the same RBF kernel. The true prior mean (the second one) is shown with a solid, black line, whereas all other prior means are shown in dashed lines.  }
    \label{fig:hillymean}
\end{figure}

Compared to Regret Balancing, which looks at the function values obtained with different priors, our algorithm looks only at the prediction errors of models obtained with different priors. As the model based on the true prior comes with predicitve performance guarantees, the sum of predicitive errors is bounded with high probability, regardless of whether the function or the search space is changing between timesteps. This makes our method applicable to nonstationary settings, where Regret Balancing cannot be used.

On high level the prior rejection mechanism might resemble the soft revision mechanism \citep{cattaneo2014continuous,augustin2021comment}. However, there is a number of important differences. Soft-revision operates on the likelihood, whereas we operate on the predictive error of the models. While linking the likelihood-based approach to guarantees on cumulative regret bounds may not be straightforward, in our proof we are able to express the regret directly in terms of the predictive error, on which our algorithm operates, allowing us to link these two notions. Soft revisions also rely on finding the prior producing the biggest marginal to decide which priors to reject, whereas in our approach the rejection is done regardless of how well other priors perform. 

As explained in the introduction, type II MLE is not always guaranteed to recover the true prior. However, it is a popular practical strategy of and as such it is interesting to compare it with the theoretically sound strategy of PE-GP-UCB. The main difference is that in PE-GP-UCB, we look at the predictive error of the model, as opposed to its likelihood, and when measuring the error for the $t$-th datapoint, we only condition the GP on $\mathcal{D}_{t-1}$, that is all observations until $t$.  In this sense, PE-GP-UCB asks retrospectively how good the models produced by a prior at each timestep $t$ are. We now empirically compare the practical strategy of MLE with the theoretically sound strategy of PE-GP-UCB and show the latter can outperform the former.

 \section{EXPERIMENTS}

 We now empirically evaluate our proposed algorithm. We conduct the evaluation on both toy and real-world problems. The first baseline we compare against is the MLE, which simply selects the prior that produces a model with the highest marginal likelihood, i.e. $p_t = \arg\max_{p \in \mathcal{U}} P(\mathcal{D}| p)$ and then optimises the UCB function $\textrm{UCB}_t^{p_t}(\bm{x}, t)$ produced by that model. Another baseline is Fully Bayesian treatment of the prior, which maximises the weighted average of UCB functions obtained under different priors with the weights being the marginal likelihood of the model with given prior, i.e. $\sum_{p \in \mathcal{U}}\textrm{UCB}_{t}^p(\bm{x}, t) P(p|\mathcal{D}_{t-1})$, where $P(p|\mathcal{D}_{t-1}) \propto P(\mathcal{D}_{t-1}| p) P(p)$ and $P(p)$ is the hyperprior (that is prior over priors). In our experiments we simply take $P(p) = \frac{1}{|\mathcal{U}|}$. The final baseline we compare against is Regret Balancing, which uses the observed function values in selecting the priors the use at future iterations. For completeness, we provide a pseudo-code in Supplementary Material Section \ref{app:rb} of the exact version of Regret Balancing that we used. Such an algorithm admits a regret bound, but only in the case, where the problem is stationary. As such, comparing the performance of Regret Balancing to PE-GP-UCB will help gauge the improvement coming from not relying on the stationarity of the objective. We detail each experiment below.  Our code is available on github\footnote{\url{https://github.com/JuliuszZiomek/PE-GP-UCB}}. To run all experiments we use one machine with AMD Ryzen 7 5800X 8-Core processor, no GPU was required.

 \begin{figure*}[ht]
    \centering
    \begin{subfigure}[b]{0.49\textwidth}
        \includegraphics[width=\textwidth]{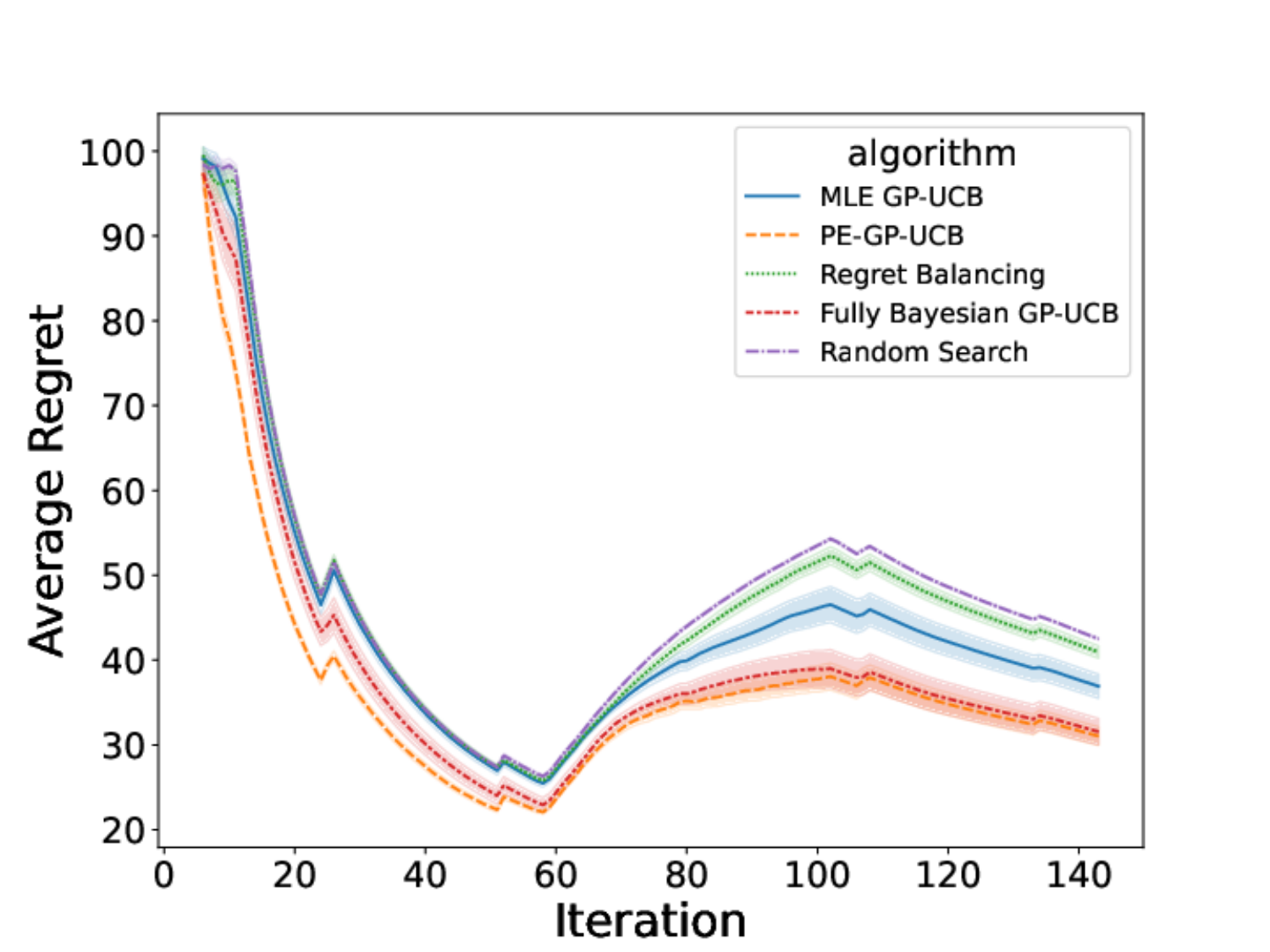}
        \caption{Regret results.}
        \label{fig:2}
    \end{subfigure}
    \hfill
    \begin{subfigure}[b]{0.49\textwidth}
        \includegraphics[width=\textwidth]{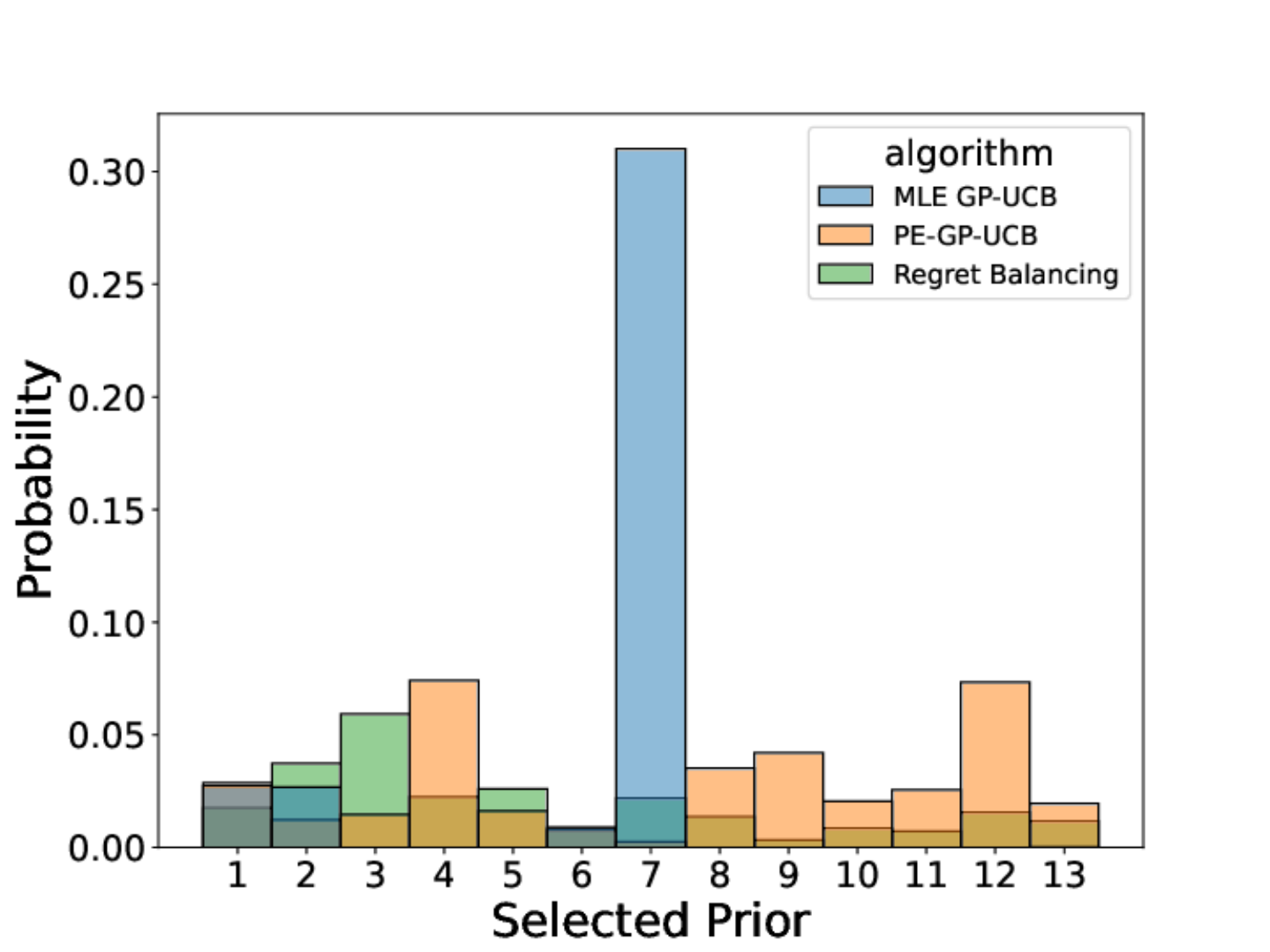}
        \caption{Histogram of prior selection frequency.}
        \label{fig:3}
    \end{subfigure}
    \caption{Results on the Intel temperature data. We ran 30 seeds and in subfigure a) we show the mean values in solid lines and standard errors as shaded regions. In subfigure b) we show the proportion of times that the methods chose each prior accross all timesteps and all seeds.}
    \label{fig:intel}
\end{figure*}

 \textbf{Toy Problem}
As the first problem, we consider a toy setup, where the ground truth function $f$ is a sample from a GP. The function $f$ is itself stationary, i.e. $\forall_{t, t^\prime \in [T]}f(\cdot, t) = f(\cdot, t^\prime)$, but the space of available points to query changes between iterations, making the problem non-stationary. 
The true GP employs an RBF kernel with a specially designed mean function that features ten distinct `hills'—regions where the mean function is higher. We use prior 2 from Figure~\ref{fig:hillymean} as the true mean function, which has one prominent hill around 0.1 and nine smaller hills.
For the estimation setup, we prepare eleven GP prior candidates, each using the same RBF kernel but with different mean functions, as shown in Figure~\ref{fig:hillymean}. Each prior features a tall hill at varying locations, along with nine smaller hills. Prior 0, however, only has ten short hills. An algorithm that can correctly identify prior 2 should achieve faster regret convergence. Furthermore, we introduce a time-varying search space 
$\mathcal{X}_t$, which alternates between the entire domain $\mathcal{X}$ and a restricted search space that excludes the true tall hill, creating an additional challenge for estimation. As such, the optimal region is excluded from the available query points at every even iteration.

We plot the cumulative regret of each algorithm in Figure \ref{fig:toy_problem}. We can see that MLE, Fully Bayesian approach, and Regret Balancing all perform similarly and outperform Random Search, however, still underperform compared to our algorithm PE-GP-UCB. The histogram in Figure \ref{fig:toy_problem} sheds some light as to why this is happening. MLE keeps selecting the same (wrong prior) and gets stuck at suboptimal values. Regret Balancing is trying out all of the priors in a roughly uniform manner, however, since the true prior is selected only a fraction of the time and half of the time the optimal region is excluded from the search space, it is likely that the algorithm will not observe better function values while using the true prior. As a result, the Regret Balancing scheme will not commit to the true prior and keep exploring indefinitely. On the other hand, our algorithm PE-GP-UCB, explores all the priors and swiftly rejects incorrect ones, committing to the true prior, resulting in the lowest regret.

 \textbf{Intel Temperature Dataset}
As the second experiment, we consider a real-world dynamic optimisation on the famous Intel Research Dataset\footnote{\url{https://db.csail.mit.edu/labdata/labdata.html}}. The dataset contains temperature measurements gathered from over 50 sensors located in the Intel Research office in Berkeley. The goal is to select the sensor with the highest temperature at each timestep and the regret is the difference between the highest and selected temperature. For the sake of optimisation, we use the data coming from 14th March 2004 and we define each timestep to be a ten-minute interval. To obtain the prior of the function, we use the previous days in March, obtaining a total of thirteen priors (one from each day). Following previous work \cite{bogunovic2016time}, for each prior we use a zero mean function and a kernel of form $k((i,t), (i^\prime,t^\prime)) = \textrm{Cov}(i,i^\prime)(1-\varepsilon)^{|t-t^\prime| / 2}$, where $\textrm{Cov}(i,i^\prime)$ is the empirical covariance between sensors $i$ and $i^\prime$ (as estimated on a given day) and $\varepsilon$ is selected to produce the best fit on that day. Since we do not know which day would be most similar to the 14th of March, we do not know which prior we should rely on. We run our method and the baseline and show the plot of average regret against time step in Figure \ref{fig:intel}. 

We can see that due to the dynamic nature of this problem, Regret Balancing scheme severely underperforms and for most iterations does not perform better than random search. MLE performs slightly better, but is still not significantly better than random search in the first half of the experiment. The Fully Bayesian approach has the advantage of not committing to any particular prior, making it more robust and outperforming MLE. However, even this strong baseline is outperformed by our algorithm PE-GP-UCB, with the difference especially visible in the beginning. Inspecting the priors selected by each method, we see that MLE commits to the prior coming from the 7th of March. At first glance, this seems sensible, as both 7th and 14th were Sundays, and we might expect that on a similar day of the week, the temperature patterns will be similar. It turns out, however, that an optimiser utilising this prior underperforms compared to PE-GP-UCB, which mostly relies on priors obtained from the 4th and 12th of March. 

\section{Conclusions}
Within this work, we proposed the first algorithm with provable regret bound, capable of solving time-varying GP bandits problem under unknown prior.
One limitation of our work is that the time-dependent part of the regret bound scales with $\sqrt{|U|}$ that is the square root of number of candidate priors. This might make the method perform poorly if the number of candidate priors is extreme. Another limitation is the aforementioned over-optimisim, which prevents the application of the proposed algorithms in certain settings. Finding a way to address those limitations consistutes an interesting direction for future work. 

\subsubsection*{Acknowledgements}
We thank anonymous reviewers who gave useful comments. Juliusz Ziomek was supported by the Oxford Ashton-Memorial Scholarship and EPSRC DTP grant EP/W524311/1. Masaki Adachi was supported by the Clarendon Fund, the Oxford Kobe Scholarship, the Watanabe Foundation, and Toyota Motor Corporation.

\bibliographystyle{aistats2024_conference}
\bibliography{example_paper}

\appendix
\onecolumn
\aistatstitle{Nonstationary Gaussian Process Bandits with Unknown Prior: 
Supplementary Materials} 

{ 
\section{Proof of Lemma \ref{lemma:noisebound}} \label{ap:noisebound}
\begin{proof}
We note that in the sum $\sum_{t \in S_T^p} \epsilon_t$, the set $S_T^p$ can depend on the noise values. However, note that the decision of which prior to choose at a given iteration $t$ (and thus whether or not $t \in S_T^p$) is made before observing the function value at that step and thus while  $\{\epsilon_t\}_{t=1}^{\tau - 1}$ can affect whether or not $\epsilon_\tau$ is included in $S_T^p$, they cannot affect its value. As such, bounding such a sum is equivalent to bounding a sum of i.i.d. variable with a history-dependent stopping rule that tells us when to stop putting more variables into the sum. We will do this by showing the inequality holds with high probability for any possible stopping time simultanously, so we can be sure that no matter what stopping time is chosen, the inequality will hold with high probability. Let $\{v_i\}_{i=1}^\infty$ be a sequence of i.i.d. random variables, each being $R$-subgaussian. We then have that:
\begin{align*}
    P\left( \exists_{t \in S_T^p}\left| \sum_{i=1}^t \epsilon_i\right| \le \sqrt{\xi_t t} \right) &= P\left(\exists_{0\le n\le T} \exists_{t \in [n]}\left| \sum_{i=1}^{t} v_i\right| \le \sqrt{\xi_t t} , |S_T^p| = n\right)\\
    &\le P\left(\exists_{0\le n\le T} \exists_{t \in [n]}\left| \sum_{i=1}^{t} v_i\right| \le \sqrt{\xi_t t}\right) \\
    & = P\left(\exists_{0 \le t \le T}\left| \sum_{i=1}^{t} v_i\right| \le \sqrt{\xi_t t}\right) \\
    &\le \sum_{t\in [T]} P\left(\left| \sum_{i=1}^{t} v_i\right| \le \sqrt{\xi_t t}\right) \\
    &\le 2\sum_{t\in [T]} \exp\left(-\frac{\xi_t}{2R^2}\right) \\
    &= 2\sum_{t\in [T]} \frac{3  \delta}{\pi^2t^2 |\mathcal{U}|} \le 2\sum_{t=1}^\infty \frac{3 \delta}{\pi^2t^2  |\mathcal{U}|} = \frac{\delta}{|\mathcal{U}|}.
\end{align*}
In the penultimate transition above we used the fact that since $v_i$ is $R$-subgaussian, for any $n > 0 $ we have:
    \begin{equation*}
        P \left( \left | \sum_{i=1}^n v_i \right| \ge \sqrt{A n} \right) \le 2 \exp \left( \frac{-A n}{2R^2 n} \right) =  2 \exp \left( \frac{-A}{2R^2} \right),
    \end{equation*}
    which is a well-known result in the literature (e.g. see Corollary 5.5 in \cite{lattimore2020bandit}). Taking union bound over priors proves the statement:
    \begin{equation*}
         P\left( \exists_{p \in \mathcal{U}}\exists_{t \in S_T^p}\left| \sum_{i=1}^t \epsilon_i\right| \le \sqrt{\xi_t t} \right) \le \sum_{p \in \mathcal{U}}\frac{\delta}{|\mathcal{U}|} = \delta .
    \end{equation*}
Note that the statement is proven for any prior $p\in \mathcal{U}$ and any $t \in S_T^p$. As such, the statement of the Lemma is a finite sample, high-probability bound and not an asymptotic statement.
\end{proof}
}

\section{Proof of Lemma \ref{lemma:maxinfogain}} \label{ap:maxinfogain}
\begin{proof}
Firstly, observe that due to Cauchy-Schwarz we have:
\begin{equation} \label{eq:cs_bound}
    \sum_{t\notin \mathcal{C}} \beta_t^+ \sigma^+_{t-1}(\bm{x}_t) \le \beta_T \sqrt{(T - |\mathcal{C}|) 
 \sum_{t\notin \mathcal{C}} (\sigma^+_{t-1})^2(\bm{x}_t)} \le \beta_T \sqrt{T 
  \sum_{p \in \mathcal{U}} \sum_{\substack{t\notin \mathcal{C} \\ p_t = p}} (\sigma^p_{t-1})^2(\bm{x}_t)} ,
\end{equation}
where $\beta_T = \max_{p \in \mathcal{U}} \beta^p_T$. We will need to introduce some new notation. For some set of inputs $A \subset \mathcal{X}$, let us define:

    \begin{equation*}
    (\sigma^u_{A})^2(\bm{x}) = k(\bm{x}, \bm{x})^u - \bm{k}^u_{A}(\bm{x})^T(K^u_{A} + \sigma^2\mathcal{I})^{-1}\bm{k}^u_{A}(\bm{x}),
    \end{equation*}
    where $K^{u^*}_A = [k^{u^*}(\bm{x},\bm{x}^\prime)]_{\bm{x},\bm{x}^\prime \in A}$ and $\bm{k}^{u^*}_A = [k^{u^*}(\bm{x},\bm{x}^\prime)]_{\bm{x}^\prime \in A}$
    Notice that under such notation $\sigma^u_{t-1} = \sigma^u_{X_{t-1}}$, where $X_{t-1} = \{\bm{x}_{i}\}_{i=1}^{t-1}$.
Let us define and $A^p_{t-1} = \{((\bm{x}_i, i), y_i): 1 \le i \le t-1 , p_t = p \}$ to be the set of points queried when $p_t = p$. For any two sets $S, S^\prime$ such that $S^\prime \subseteq S$ we have $\sigma^p_{S^\prime}(\bm{x}) \le \sigma^p_{S}(\bm{x})$ for all $\bm{x} \in \mathcal{X}$. As $A^p_{t-1} \subseteq \mathcal{D}_{t-1}$, we have:
\begin{equation*}
     (\sigma^+_{t-1})^2(\bm{x}_t) =  (\sigma^p_{\mathcal{D}_{t-1}})^2(\bm{x}_t) \le  (\sigma^{p}_{A^p_{t-1}})^2(\bm{x}_t)  \le C \log (1 + R^{-2}( \sigma^p_{A_{t-1}^p}(\bm{x}))^2 ),
\end{equation*}
 for some $C$ (depending on $R$), where the last inequality is true due to the same reasoning as in the proof of Lemma 5.4 in \cite{srinivas2009gaussian}. Applying this fact to Inequality \ref{eq:cs_bound} gives:
\begin{equation*}
    \beta_T \sqrt{T 
  \sum_{p \in \mathcal{U}} \sum_{\substack{t\notin \mathcal{C} \\ u_t = p}} (\sigma^p_{t-1})^2(\bm{x}_t)} \le \beta_T \sqrt{CT 
  \sum_{p \in \mathcal{U}} \sum_{\substack{t\notin \mathcal{C} \\ u_t = p}} \log (1 + R^{-2}( \sigma^p_{A_{t-1}^p}(\bm{x}))^2 )} 
\end{equation*}
\begin{equation*}
    = \beta_T \sqrt{CT 
  \sum_{p \in \mathcal{U}} \gamma_{|A^p_{t-1}|}^{p}} \le \beta_T \sqrt{CT |\mathcal{U}|  \gamma_{T}}
\end{equation*}
where the penultimate equality follows from Lemma 5.3 from \cite{srinivas2009gaussian}.
\end{proof}

{
\section{Proof of Lemma \ref{lemma:bayesianfunctionbound}} \label{ap:bayesianfunctionbound_proof}
\begin{proof}

    We will consider two cases, depending on cardinality of $|\mathcal{X}|$. 

\underline{\textbf{If $|\mathcal{X}| < \infty$}}

Using the same proof idea as Lemma 5.1. of \cite{srinivas2009gaussian}, where we simply replace the posterior mean $\mu_{t-1}$ and variance $\sigma^2_{t-1}$ with the prior mean $\mu$ and variance $\sigma$ we get that with probability at least $1-\delta_c$, for all $\bm{x} \in \mathcal{X}$ and all $t^\prime \in [T]$:
\begin{equation*}
    |f(\bm{x}, t^\prime) - \mu^{p^\star}(\bm{x}, t^\prime)|  \le \beta^{p^\star}_t \sigma^{p^\star}_{0}(\bm{x}, t^\prime) \le \beta^{p^\star}_t k^{p^\star}((\bm{x}, t^\prime), (\bm{x}, t^\prime)) \le \hat{\beta}_t^{p^\star}, 
\end{equation*}
where $\beta^{p}_t = \sqrt{2\log\left( \frac{|\mathcal{X}| \pi^2 t^2}{2\delta_c} \right)}$ As such:
\begin{equation*}
   \max_{t^\prime \in [t]} \max_{\bm{x} \in \mathcal{X}} | f(\bm{x}, t^\prime)| \le \beta^{p^\star}_t + \max_{t^\prime \in [t]}\max_{\bm{x} \in \mathcal{X}}|\mu^{p^\star}(\bm{x}, t^\prime)| \le B_t.
\end{equation*}

\underline{\textbf{If $|\mathcal{X}| = \infty$}}

Using the same proof idea as Appendix C.1. of \cite{bogunovic2016time}, where we simply replace the posterior mean $\mu_{t-1}$ and variance $\sigma^2_{t-1}$ with the prior mean $\mu$ and variance $\sigma$ we get that with probability at least $1-\delta_c$, for all $\bm{x} \in \mathcal{X}$ and all $t^\prime \in [T]$:
\begin{equation*}
      |f(\bm{x}, t^\prime) - \mu^{p^\star}([\bm{x}]_1, t^\prime)  \le \beta^{p^\star}_{t^\prime} \sigma^{p^\star}([\bm{x}]_1, t^\prime) + 1 \le \beta^{p^\star}_{t^\prime} k^{p^\star}([\bm{x}]_1,[\bm{x}]_1, t^\prime) + 1 \le \beta^{p^\star}_{t^\prime} + 1 ,
\end{equation*}
where $\beta_t^{p} =  \sqrt{2 \log\left(\frac{3\pi^2 t^2}{2\delta_C}\right) + 2d \log\left( dt^2rb^p\sqrt{\log\left(\frac{3da^p}{2\delta_C}\right)}\right)}
$. As such:
\begin{equation*}
    \max_{t^\prime \in [t]}\sup_{\bm{x} \in \mathcal{X}} | f(\bm{x}, t^\prime)| \le \beta^{p^\star}_{t^\prime} +  \max_{t^\prime \in [t]} \max_{\bm{x} \in D_t}|\mu^{p^\star}(\bm{x}, t^\prime)| + 1 \le \beta^{p^\star}_{t^\prime} +  \max_{t^\prime \in [t]} \sup_{\bm{x} \in \mathcal{X}}|\mu^{p^\star}(\bm{x}, t^\prime)| + 1 \le B_t. 
\end{equation*}

\end{proof}
}

\section{Regret Balancing for Unknown Prior} \label{app:rb}
For completeness, below we provide a version for Regret Balancing for the case of unknown prior, adapted from \cite{ziomek2024bayesian}, who derived regret bound for these kinds of algorithms in BO, but only if the underlying function is not changing with time. In our experiments, we set $\{\xi_t\}_{t=1}^T$ and
$\{\beta^u_t\}_{t=1}^T$ in the same way as for PE-GP-UCB. We also set the suspected regret bounds to simply $R^p(t) = t$ for all $p \in \mathcal{U}$, as in the first experiment (toy problem) the priors differ only with respected to mean (and thus the well-specified regret bound is the same under all priors) and in the case second experiment (Intel data), we do not know the exact regret bound scaling due to the shape of used kernel. 

\begin{algorithm}
\caption{Regret Balancing for Unknown Prior}\label{alg:hb_bo}
\begin{algorithmic}[1]
\REQUIRE  suspected regret bounds $R^p(\cdot)$; confidence parameters $\{\xi_t\}_{t=1}^T$ and
$\{\beta^u_t\}_{t=1}^T$ 

\STATE Set $\mathcal{D}_{0} = \emptyset$, $\mathcal{U}_1 = \mathcal{U}_0$ ,$S^p_0 = \emptyset$ for all $p \in \mathcal{U}_1$,  
\FOR{$t = 1, \dots, T$}
\STATE Select prior  $u_t = \arg\min_{p \in \mathcal{U}_t} R^p(|S_{t-1}^{p}| + 1)$
\STATE Select point to query $\bm{x}_t =  \underset{\bm{x} \in \mathcal{X}}{\arg\max} \textrm{ UCB}_{t-1}^{p_t}(\bm{x})$
\STATE Query the black-box $y_t = f(\bm{x}_t, t)$ 
\STATE Update data buffer $\mathcal{D}_t = \mathcal{D}_{t-1} \cup ((x_t, t), y_t)$
\STATE For each $p \in \mathcal{U}_{t}$, set $S_t^{p} = \{\tau  = 1,\dots, t: p_{\tau} = p\}$
\STATE Initialise prior set for new iteration $\mathcal{U}_{t+1} := \mathcal{U}_{t}$
\IF{ $\forall_{p \in \mathcal{U}_t} |S_t^p| \neq 0$}
\STATE Define $L_t(p) = \left( \frac{1}{|S_t^{p}|} \sum_{\tau \in S_t^{p}}y_{\tau} - \sqrt{\frac{\xi_{t}}{|S_{t}^{p}|}} \right)$

\STATE $\mathcal{U}_{t+1} = \left\{ p \in \mathcal{U}_t: L_t(u) +  \frac{1}{|S_t^{p}|} \sum_{\tau \in S_t^p} \beta_\tau^p \sigma^p_{\tau-1}(\bm{x}_\tau)  \ge  \max_{p^\prime \in \mathcal{U}_{t}} L_t(p^\prime) \right\}$

\ENDIF

\ENDFOR

\end{algorithmic}
\end{algorithm}

{
\section{Proof of Proposition \ref{corollary:win}} 
Using the same reasoning as in Appendix E of \cite{bogunovic2016time} we have that 
in squared exponential case $\gamma_T = \Tilde{\mathcal{O}}(\max\{T \varepsilon^{1/3}, 1\})$ and in $\nu$-Matern case $\gamma_T = \Tilde{\mathcal{O}}( \max\{T \varepsilon^{\frac{1-c}{3-c}} , T^c\})$ with $c = \frac{d(d+1)}{2\nu + d(d+1)}$.

\section{Proof of Lemma \ref{lemma:tstar}}
To prove the first statement we need to show that with probability 1, we have $\forall_{\epsilon > 0} \exists_{T^\prime(\epsilon)>0} \forall_{T > T^\prime(\epsilon)} \frac{t^\star}{T^\prime(\epsilon)} < \epsilon$.
 Note that since confidence parameters $\xi_t$ and $\beta_t$ do not require the knowledge of $T$ beforehand, we can always just decide to continue run the algorithm for more iterations after exceeding some inital iteration number. Let $\tau_i$ be the iteration, where we reject the $i$th prior if we run the algorithm for at least $\tau_i$ iterations. Note that $(\tau_1, \dots, \tau_{|\mathcal{U}|})$ are random variables that depend on the randomness in sampling black-box $f$ from the prior (and its random evolution through time), and on the randomness in noise $\{\epsilon_t\}_{t=1}^\infty$, but do not depend on any randomness in the algorithm. Thus conditioning on noise variables and black-box function and its evolution, we have that $(\tau_1, \dots, \tau_{|\mathcal{U}|})$ are determinstic variables. Let us denote $\tau_i = \infty$ if less than $i$ priors were rejected. Thus $t^\star \le  \max_{i=1,\dots,|\mathcal{U}|} \{\tau_i| \tau_i \neq \infty\}$ and conditioned on the aforementioned randomness $t^\star$ also becomes deterministic. As such, setting $T^\prime(\epsilon) = \frac{2 \max_{i=1,\dots,|\mathcal{U}|} \{\tau_i| \tau_i \neq \infty\}}{\epsilon}$, we get that:
 \begin{equation*}
     P\left( \forall_{T > T^\prime(\epsilon)} \frac{t^\star}{T} \le \frac{\max_{i=1,\dots,|\mathcal{U}|} \{\tau_i| \tau_i \neq \infty\}}{T} < \epsilon \Big | \{f(x_t, t)\}_{t=1}^\infty, \{\epsilon_t\}_{t=1}^\infty \right) = 1 ,
 \end{equation*}
 and as this construction can be done for any $\epsilon > 0$, we have:
 \begin{equation*}
     P\left( \forall_{\epsilon > 0}\forall_{T > T^\prime(\epsilon)} \frac{t^\star}{T} < \epsilon \Big | \{f(x_t, t)\}_{t=1}^\infty, \{\epsilon_t\}_{t=1}^\infty \right) = 1 .
 \end{equation*}
 Thus by law of total probability $P(A) = \int P(A|B = b)p(b)db$, we get that:
 \begin{equation*}
     P\left(\forall_{\epsilon > 0} \forall_{T > T^\prime(\epsilon)} \frac{t^\star}{T} < \epsilon  \right) = 1.
 \end{equation*}
 This finishes the proof for the first statement. To prove the second statement, we can use the same reasoning as above and set $T^\prime(\epsilon) = \frac{2B_{t^\star}}{\epsilon}$, as since $t^\star$ is deterministic, so is $B_{t^\star}$

}
\section*{Checklist}

 \begin{enumerate}

 \item For all models and algorithms presented, check if you include:
 \begin{enumerate}
   \item A clear description of the mathematical setting, assumptions, algorithm, and/or model. [Yes]
   \item An analysis of the properties and complexity (time, space, sample size) of any algorithm. [Yes (regret analysis)]
   \item (Optional) Anonymized source code, with specification of all dependencies, including external libraries. [Yes]
 \end{enumerate}

 \item For any theoretical claim, check if you include:
 \begin{enumerate}
   \item Statements of the full set of assumptions of all theoretical results. [Yes]
   \item Complete proofs of all theoretical results. [Yes]
   \item Clear explanations of any assumptions. [Yes]     
 \end{enumerate}

 \item For all figures and tables that present empirical results, check if you include:
 \begin{enumerate}
   \item The code, data, and instructions needed to reproduce the main experimental results (either in the supplemental material or as a URL). [Yes]
   \item All the training details (e.g., data splits, hyperparameters, how they were chosen). [Yes]
         \item A clear definition of the specific measure or statistics and error bars (e.g., with respect to the random seed after running experiments multiple times). [Yes]
         \item A description of the computing infrastructure used. (e.g., type of GPUs, internal cluster, or cloud provider). [Yes]
 \end{enumerate}

 \item If you are using existing assets (e.g., code, data, models) or curating/releasing new assets, check if you include:
 \begin{enumerate}
   \item Citations of the creator If your work uses existing assets. [Not Applicable]
   \item The license information of the assets, if applicable. [Not Applicable]
   \item New assets either in the supplemental material or as a URL, if applicable. [Not Applicable]
   \item Information about consent from data providers/curators. [Not Applicable]
   \item Discussion of sensible content if applicable, e.g., personally identifiable information or offensive content. [Not Applicable]
 \end{enumerate}

 \item If you used crowdsourcing or conducted research with human subjects, check if you include:
 \begin{enumerate}
   \item The full text of instructions given to participants and screenshots. [Not Applicable]
   \item Descriptions of potential participant risks, with links to Institutional Review Board (IRB) approvals if applicable. [Not Applicable]
   \item The estimated hourly wage paid to participants and the total amount spent on participant compensation. [Not Applicable]
 \end{enumerate}

 \end{enumerate}

\end{document}